\documentclass{bmvc2k}

\title{Global Proxy-based Hard Mining for \\Visual Place Recognition}

\addauthor{Amar Ali-bey}{amar.ali-bey.1@ulaval.ca}{1}
\addauthor{Brahim Chaib-draa}{brahim.chaib-draa@ift.ulaval.ca}{1}
\addauthor{Philippe Gigu\`ere}{philippe.giguere@ift.ulaval.ca}{1}

\addinstitution{
 Department of Computer Science and Software Engineering\\
 Universit\'e Laval\\
 Qu\'ebec City, Canada
}

\runninghead{Ali-bey \etal}{Global Proxy-based Hard Mining for VPR}

\def\etal{\emph{et al}\bmvaOneDot}

\usepackage{geometry}
\usepackage[normalem]{ulem}
\usepackage{multirow}
\usepackage{graphicx}
\usepackage{booktabs}
\usepackage{pifont}
\usepackage{amsfonts}
\usepackage[ruled,vlined]{algorithm2e}
\usepackage{algpseudocode}
\usepackage[bookmarks=false]{hyperref}
\begin{document}

\maketitle

\begin{abstract}
Learning deep representations for visual place recognition is commonly performed using pairwise or triple loss functions that highly depend on the hardness of the examples sampled at each training iteration. Existing techniques address this by using computationally and memory expensive offline hard mining, which consists of identifying, at each iteration, the hardest samples from the training set. In this paper we introduce a new technique that performs global hard mini-batch sampling based on proxies. To do so, we add a new end-to-end trainable branch to the network, which generates efficient place descriptors (one proxy for each place). These proxy representations are thus used to construct a global index that encompasses the similarities between all places in the dataset, allowing for highly informative mini-batch sampling at each training iteration. Our method can be used in combination with all existing pairwise and triplet loss functions with negligible additional memory and computation cost. We run extensive ablation studies and show that our technique brings new state-of-the-art performance on multiple large-scale benchmarks such as Pittsburgh, Mapillary-SLS and SPED. In particular, our method provides more than 100\% relative improvement on the challenging Nordland dataset. Our code is available at \url{https://github.com/amaralibey/GPM}
\end{abstract}

\section{Introduction}
\label{sec:intro}
\begin{figure}[th]
\centering
  \includegraphics[width=\linewidth]{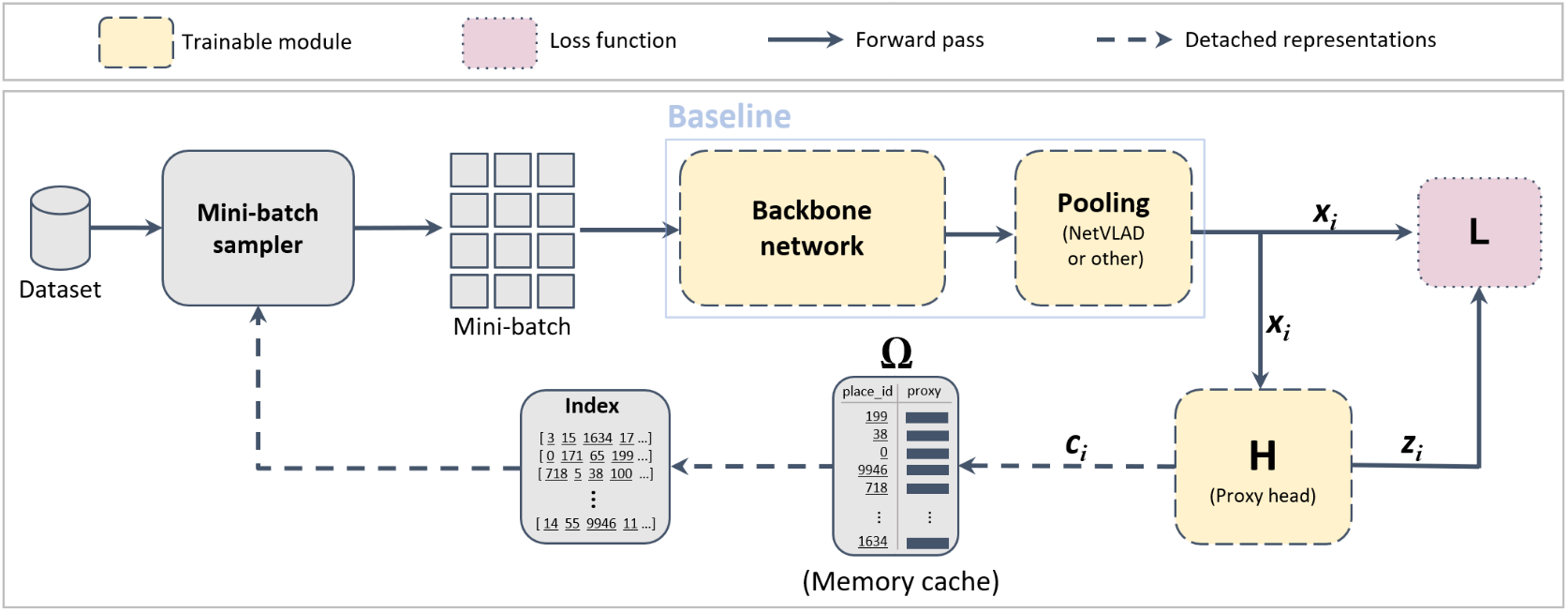}
  \vspace{0.5pt}
  \caption{A diagram of our proposed method. We add a new end-to-end trainable branch to the network (proxy head $\mathcal{H}$) that projects highly dimensional vectors $\mathbf{x}_i$ into very compact representations $\mathbf{z}_i$ ; we use the latter to compute one proxy descriptor $\mathbf{c}_i$ for each place in the mini-batch. We detach each proxy from the computation graph and cache it into a memory bank $\Omega$. Then, at the begining of each epoch, we construct an index upon $\Omega$, in which places are gathered together according to the similarity of their proxies. This index is used to sample mini-batches containing similar places, which yields highly informative pairs or triplets. We call this strategy Global Proxy-based Hard Mining (GPM).}
\label{fig:arch}
\end{figure}
Visual place recognition (VPR) consists of determining the location of a place depicted in a query image by comparing it to a database of previously visited places with known geo-references. This is of major importance for many robotics and computer vision tasks, such as autonomous driving~\cite{chowdhary2013gps, maddern20171}, SLAM~\cite{milford2012seqslam, engel2014lsd}, image geo-localization~\cite{baik2020domain, hausler2021patch, wang2022transvpr} and 3D reconstruction~\cite{cieslewski2016point, sattler2017large}.
Recently, advances in deep learning~\cite{menghani2021efficient} have made retrieval-based place recognition a preferable choice for efficient and large-scale localization. Current VPR techniques \cite{arandjelovic2016netvlad, liu2019stochastic, warburg2020mapillary, thoma2020soft, zhu2020regional, hausler2021patch, wang2022transvpr} use metric learning loss functions to train deep neural networks for VPR. These loss functions operate on the relationships between images in a mini-batch. As such, representations of images from the same place are brought closer and those from different places are distanced~\cite{musgrave2020metric}. For instance, in the most used architecture for VPR, NetVLAD~\cite{arandjelovic2016netvlad, liu2019stochastic, warburg2020mapillary, hausler2021patch, wang2022transvpr}, the network is trained using a triplet ranking loss function that operates on triplets, each of which consists of a query image, a positive image depicting the same place as the query, and a negative image depicting a different place. Moreover, the triples need to be informative in order for the network to converge~\cite{hermans2017defense}, meaning that for each query, the negative must be hard for the network to distinguish from the positive. To do so, these techniques rely on offline hard negative mining, where every image representation generated by the network is kept in a memory bank (cache), to be used offline (out of the training loop) to find the hardest negatives for each training query. Although offline mining allows the network to converge~\cite{warburg2020mapillary}, it involves a large memory footprint and computational overhead.
Another approach for informative example mining is online hard negative mining (OHM)~\cite{hermans2017defense, wu2017sampling}, which consists of first forming mini-batches, by randomly selecting a subset of places from the dataset and sampling images from each of them. Then, in a later stage of the forward pass, select only the most informative triples (or pairs) present in the mini-batch and use them to compute the loss. Nevertheless, randomly constructed mini-batches can generate a large number of triplets (or pairs), most of which may be uninformative~\cite{hermans2017defense}. Yet selecting informative samples is crucial to robust feature learning~\cite{musgrave2020metric}. The advantage of OHM is that there is no memory bank (cache) and no out-of-the-loop mining step. However, as training progresses and the network eventually learns robust representations, the fraction of informative triplets (or pairs) within the randomly sampled mini-batches becomes limited (i.e., the network becomes good at distinguishing hard negatives). Therefore, it's recommended to use very large batch sizes~\cite{hermans2017defense} to potentially increase the presence of hard examples at each iteration.

In this work, we propose a new globally informed mini-batch sampling technique, which instead of randomly sampling places at each iteration, it uses a proxy index to construct mini-batches containing visually similar places. The main idea behind our technique is the following: instead of caching highly dimensional individual image descriptors to mine hard negatives, we propose to add an auxiliary branch that computes compact place-specific representations that we call proxies. Thus, each place in the dataset can be globally represented by one low-dimensional proxy that can be effectively cached during the training. This allows us to build an index in which places are gathered in the same mini-batch according to the similarity of their proxies. Our technique involves negligible computational and memory overhead, while drastically improving performance.

\section{Related Work}
\label{sec:related}
\subsection{Visual Place Recognition}\label{ssec:vpr}
Most state-of-the-art techniques in VPR~\cite{arandjelovic2016netvlad, liu2019stochastic, seymour2019semantically, warburg2020mapillary, kim2017learned, liu2020digging, hausler2021patch, wang2022transvpr} train the network with mini-batches of triplets of images. Such techniques employ offline hard negative mining to form informative triplets. This is done by storing in a memory cache all image representations generated during the training, and using $k$-NN to retrieve, for each training query, the hardest negatives among all references in the cache and form informative triplets (the hard negatives are the images that do not depict the same place as the query but are too close to it in the representation space). However, most SOTA methods generate highly dimensional representations during the training phase, for instance, techniques that rely on NetVLAD~\cite{arandjelovic2016netvlad} generate descriptors of size $d = 32768$. As a result, caching representations when training with large datasets such as Mapillary~SLS~\cite{warburg2020mapillary} or GSV-Cities~\cite{ali2022gsv} quickly becomes infeasible, because of both the computational overhead and the memory footprint of $k$-NN, which has a computational complexity of $\mathcal{O}(QRd)$ and a memory footprint of $\mathcal{O}(Rd)$~\cite{cunningham2021k}, where $R$ is the number of reference samples (cached representations), $d$ the dimensionality of each sample, and $Q$ is the number of queries to be searched.
In \cite{thoma2020soft, arandjelovic2016netvlad, liu2019stochastic} the representations of all the training examples of Pitt250k dataset are cached. Then, after a fixed number of iterations, the training is paused and the cache is used to mine the hardest $10$ negatives for each training query (to form hard triplets). Importantly, the cache is recalculated every $250$ to $1000$ iterations. Warburg \etal~\cite{warburg2020mapillary} trained NetVLAD on Mapillary-SLS, which is a dataset comprising $1.6$M images. Faced with the huge memory overhead, they used a subcaching strategy, where only a subset of the training images are cached, from which the hard negatives were periodically mined. Note that, if the NetVLAD representations of all images in MSLS dataset~\cite{warburg2020mapillary} were cached, the memory cache would be $196$GB in size.
From the above, it is evident that the extra memory and computational cost of offline hard mining for VPR remains an issue to be addressed. 

\subsection{Deep Metric Learning}\label{ssec:dml}
Place recognition networks are generally trained using ranking loss functions issued from deep metric learning~\cite{zhang2021visual}, such as triplet ranking loss~\cite{schroff2015facenet} and contrastive loss~\cite{thoma2020soft}. However, during the training, deep metric learning (DML) networks often generate very compact representations compared to VPR, ranging from $d =128$ to $d=512$~\cite{chen2021deep}. This makes any caching mechanism much less greedy and computationally inexpensive. Related to our work are DML approaches~\cite{ge2018deep, smirnov2018hard} that perform negative mining on class-level representations (a class could be regarded as the equivalent of a place in VPR), under the assumption that class-level similarity is a good approximation of the similarity between instances.  Smirnov~\etal~\cite{ge2018deep} developed a technique that constructs a hierarchical tree for the triplet loss function. The strategy behind their approach is to store class-level representations during the training, identify neighbouring classes and put them in the same mini-batch, resulting in more informative mini-batches that can be further exploited by online hard mining. Applying these techniques directly to train VPR networks would require to cache highly dimensional image-level representations (e.g. $32$K for NetVLAD), which is not feasible when the training dataset contains thousands of different places.

\section{Methodology}
\label{sec:method}
As mentioned above, VPR techniques generate highly dimensional representations, making caching and hard mining with $k$-NN impractical for large-scale datasets. Knowing that the complexity of $k$-NN is linearly  dependent on the number of references $Q$ that need to be cached and their dimensionality $d$~\cite{cunningham2021k}. And considering that the only purpose of the caching mechanism is to help retrieve hard examples. We propose to project the highly dimensional pooling representations (e.g. the resulting NetVLAD representations) into a separate branch ($\mathcal{H}$ in figure~\ref{fig:arch}) that we call \textit{proxy head}. $\mathcal{H}$ is an end-to-end trainable module that learns place-specific compact vectors of significantly smaller dimension compared to the pooling module. During each epoch, we capture and cache the semantics of each place (instead of each image) with one compact vector, acting as its global proxy. Therefore, the number of proxies to be cached is one order of magnitude smaller than the number of images in the dataset (considering that a place is generally depicted by $8$ to $20$ images as in GSV-Cities~\cite{ali2022gsv}). Most importantly, we can choose $d'$ the dimensionality of the proxy head $\mathcal{H}$ to be several orders of magnitude smaller than $d$ the dimensionality of the pooling layer. This allows to perform global hard mining based on the compact-proxies, with negligible additional memory and computation cost as we show in section~\ref{sec:exp} (i.e., using $k$-NN on the proxies is orders of magnitude more efficient).

\subsection{Representation Learning for VPR}
Given a dataset of places $\mathcal{D} = \left\{P_1, P_2, ..., P_N\right\}$ where $P_i = \left( \left\{I_1^i, I_2^i, ..., I_{|P_i|}^i\right\}, y_i \right)$ is a set of images depicting the same place and sharing the same identity (or label) $y_i$. The goal is to learn a function $\mathit{f_{\mathbf{\theta}}}$ which is, in most cases, a deep neural network composed of a backbone network followed by a pooling layer (e.g., NetVLAD). The network $\mathit{f_{\mathbf{\theta}}}$ takes an input image $I_i$ and outputs a representation vector $\mathbf{x}_i \in \mathbb{R}^{d}$ such that the similarity of a pair of instances $\left(\mathbf{x}_i, \mathbf{x}_j\right)$ is higher if they represent the same place, and lower otherwise.

As the generated representation $\mathit{f_{\theta}}\left(I_i\right) = \mathbf{x}_i$ is highly dimensional (i.e., $d = 32$k for NetVLAD~\cite{arandjelovic2016netvlad}), we propose to project it further in a separate branch of the network, that we call \textit{proxy head} ($\mathcal{H}$), represented by a function $\mathit{h_{\mathbf{\psi}}} : \mathbb{R}^{d} \mapsto \mathbb{R}^{d'}$ and projects the outputs from the pooling layer to a smaller Euclidean space where $d' << d$ as illustrated in figure~\ref{fig:arch}. Formally, for each  vector $\mathbf{x}_i$, the proxy head produces a compact projection $\mathbf{z}_i$ as follow:
\begin{equation}\label{eq1}
    \mathbf{z}_i  = \mathit{h_{\mathbf{\psi}}} \left(  \mathit{f_{\theta}}\left(I_i\right) \right) 
     =  \mathit{h_{\mathbf{\psi}}} \left(   \mathbf{x}_i \right)
\end{equation}
In this work, $\mathcal{H}$ is a fully connected layer that projects $d$-dimensional inputs to $d'$-dimensional outputs followed by $L2$ normalization. This gives us  the control of the proxy dimensionality $d'$. However, $\mathcal{H}$ could also be an MLP or a trainable module of different architecture.
We use backpropagation to jointly learn the parameters $\mathbf{\theta}$ and $\mathbf{\psi}$, using pair based (or triplet based) loss functions from metric learning literature~\cite{musgrave2020metric} such as Contrastive loss~\cite{hadsell2006dimensionality}, Triplet loss~\cite{hermans2017defense} and Multi-Similarity loss~\cite{wang2019multi}. \textbf{Note}: since the proxy head is only used during the training phase (to mine hard samples) and discarded during evaluation and test, we might not need to backpropagate the gradient from $\mathcal{H}$ back to the pooling layer. Quantitative experiments show that this does not affect performance.

\subsection{Global Proxy-based Hard Mining (GPM)}\label{ssec:gpm}
Traditionally, during the training phase, each mini-batch is formed by randomly sampling $M$ places from the dataset, then picking $K$ images from each one of them, thus resulting in a mini-batch $\mathcal{B}$ of size $M \times K$.
The goal of global hard hard mining is to populate each training mini-batch with $M$ similar places, which in turn yields hard pairs and triplets, potentially inducing a higher loss value, thereby learning robust and discriminative representations. For this purpose, we use the representations generated by the proxy head $\mathcal{H}$, and compute for each place $P_i \in \mathcal{B}$, a single compact descriptor $\mathbf{c}_i$ as follows:
\begin{equation}
    \mathbf{c}_i = \frac{1}{|P_i|} \sum_{I \in P_i} \mathit{h_{\mathbf{\psi}}} \left(   \mathit{f_{\theta}}\left(I\right) \right) 
\end{equation}
where $\mathbf{c}_i$ corresponds to the average of the proxy representations of the images depicting $P_i$ in the mini-batch $\mathcal{B}$. During the training we regard $\mathbf{c}_i$ as a global descriptor (a proxy) of $P_i$ and cache it along with its identity $y_i$ into a memory bank $\Omega$. Then, at the end of each epoch, we use $k$-NN to build an index upon $\Omega$, in which places are gathered together according to the similarity of their proxies (similar places need to appear in the same mini-batch) as in Algorithm~\ref{algo_index}.

\begin{algorithm}
\SetKwInOut{Input}{input}
\SetKwInOut{Output}{output}
\Input{$\Omega$: the memory bank comprising proxies representing all places in the dataset \\ $M$: the number of places per mini-batch.}
\Output{$\mathcal{L}$: a list of tuples, where each tuple contains $M$ identities of places that need to be sampled in the same mini-batch.}
\BlankLine
\nl $\mathcal{S} \leftarrow k\text{-NN}(k=M)$ \Comment{Initialize a $k$-NN module $\mathcal{S}$ with $k$ equal to $M$ the number of places per mini-batch.}

\nl $\mathcal{S}\text{.add}(\Omega)$ \Comment{Add the contents of $\Omega$ to $\mathcal{S}$ as references.}

\While{$\mathcal{S} \neq \emptyset$}{
\nl Randomly pick a place $c_i$ from $\mathcal{S}$

\nl $\mathbf{T} \leftarrow  \mathcal{S}\text{.search}(c_i)$ \Comment{Search $\mathcal{S}$ for the $M$-most similar places to $c_i$.}

\nl $\mathcal{L} \leftarrow \mathcal{L}\cup \mathbf{T}$ \Comment{Append the $M$ identities to $\mathcal{L}$.}

\nl $\mathcal{S} \leftarrow \mathcal{S} \setminus \mathbf{T}$ \Comment{Remove from $\mathcal{S}$ all places present in $\mathbf{T}$.}
}
\caption{Index based mini-batch sampling}\label{algo_index}
\end{algorithm}

For the epoch that follows, the mini-batch sampler picks one tuple from $\mathcal{L}$ at each iteration, yielding in $M$ similar places. We then pick $K$ images from each place resulting in highly informative mini-batches of size $M \times K$. Qualitative results in section~\ref{ssec:qualitative} show the effectiveness of our approach in constructing informative mini-batches. 

\vspace{5pt}
\noindent\textbf{Connection to proxy-based loss functions.} Deep metric learning techniques that employ the term ‘\textit{proxy}’, such as \cite{kim2020proxy, yang2022hierarchical, yao2022pcl}, are fundamentally different from our approach, in that, they learn proxies at the loss level, and optimize on the similarity between the proxies and individual samples in the mini-batch. However, learning proxies at the loss level forces them to be of the same dimensionality as the individual samples (e.g., $32$K if used to train NetVLAD). In contrast, we learn compact proxies independently of the loss function, and use them only to construct informative mini-batches.
\section{Experiments}\label{sec:exp}
\textbf{Dataset and Metrics.} GSV-Cities dataset~\cite{ali2022gsv} is used for training, it contains $65$k different places spread on numerous cities around the world, totalling $552$k images. For testing, we use the following $4$ benchmarks,  Pitts250k-test~\cite{torii2013visual}, MSLS~\cite{warburg2020mapillary}, SPED~\cite{zaffar2021vpr} and Nordland~\cite{zaffar2021vpr} which contain, respectively, $8$K, $750$, $607$ and $1622$ query images, and $83$k, $19$k, $607$ and $1622$ reference images. We follow the same evaluation metric as~\cite{arandjelovic2016netvlad, warburg2020mapillary, zaffar2021vpr} where the recall@K is reported.

\noindent\textbf{Default Settings.}
In all experiments, we use ResNet-50\cite{he2016deep} as backbone network, pretrained on ImageNet~\cite{krizhevsky2012imagenet} and cropped at the last residual bloc; coupled with NetVLAD~\cite{arandjelovic2016netvlad} as a pooling layer, we chose NetVLAD because it's the most widely used pooling technique that showed consistent SOTA performance. Stochastic gradient descent (SGD) is utilized for optimization, with momentum $0.95$ and weight decay $0.0001$. The initial learning rate on $0.05$ is multiplied by $0.3$ after each $5$ epochs. We train for a maximum of $30$ epochs using images resized to $224\times 224$.
Unless otherwise specified, we use mini-batch containing $M=60$ places, each of which depicted by $K=4$ images ($240$ in total) and fix the output size of the proxy head $d'$ to $128$ when applicable.

\subsection{Effectiveness of GPM}
To demonstrate the effectiveness of out proposed method, we conduct ablation studies on $4$ different VPR benchmarks. We illustrate the effect of using our technique (GPM) alongside three different loss functions, namely, Contrastive loss \cite{hadsell2006dimensionality}, Triplet loss \cite{hermans2017defense} and Multi-Similarity loss~\cite{wang2019multi}. 
For each loss function, we conducted four  test scenarios (one on each line) as shown in Table~\ref{tab:my-table}. First, we train the network with randomly constructed batches without OHM or GPM (baseline \#1). In the second scenario, we add GPM to the first baseline and show the effect of globally informed sampling provided by our method. The results demonstrate that GPM alone can greatly improve performance of all three loss functions. For example, the triplet loss improved recall@1 (in absolute value) by $4.3, 4.1, 3.6$ and $3.4$ points on Pitts250k, MSLS, SPED and Nordland respectively, while Multi-Similarity loss improved by $5.4, 8.5, 13.9$ and $8.6$ points.  

In the third scenario (baseline \#2), online hard mining (OHM) is used during the training without GPM. This consists of selecting the most informative pairs or triplets from randomly sampled mini-batches. The results show that OHM can improve performance over baseline~\#1, which is consistent with the existing literature~\cite{hermans2017defense}. 

For the last scenario, we used GPM combined with baseline \#2 (i.e., mini-batches are sampled using GPM and then further exploited by OHM), results show that our technique (GPM) consistently outperform the baseline. For instance, contrastive loss improved recall@1 (in percentage points) by $5.9$ on Pitts250k, $4.7$ on MSLS, $10.1$ on SPED and $16.8$ on Nordland. Note that the relative performance boost introduced by GPM on Nordland is more than $100\%$ for both contrastive and triplet loss. The best overall performance is achieved using Multi-Similarity loss which boosted the recall@1 over baseline~\#2 by, respectively, $2.0, 4.6, 4.8$ and $9.4$ points on the four benchmarks. This ablation study highlights the effectiveness of GPM compared to randomly constructed mini-batches.

\begin{table}
\centering
\resizebox{\textwidth}{!}{%
\begin{tabular}{l||cc|ccc|ccc|ccc|ccc}
\multirow{2}{*}{Loss function}    & \multicolumn{2}{c|}{Hard mining} & \multicolumn{3}{c|}{Pitts250k-test} & \multicolumn{3}{c|}{MSLS-val} & \multicolumn{3}{c|}{SPED} & \multicolumn{3}{c}{Nordland} \\ \cline{2-15} 
                                  & OHM             & GPM            & R@1        & R@5        & R@10      & R@1      & R@5      & R@10    & R@1     & R@5    & R@10   & R@1      & R@5     & R@10    \\ \hline\hline
\multirow{4}{*}{Triplet}          &                 &                & 77.0       & 90.0       & 93.6      & 67.7     & 79.2     & 82.4    & 53.7    & 69.5   & 75.8   & 8.4      & 16.3    & 20.6    \\
                                  &                 & \ding{51}      & 81.3       & 91.9       & 94.9      & 71.8     & 82.0     & 86.3    & 57.3    & 71.8   & 77.8   & 11.8     & 20.3    & 25.9    \\ \cline{2-15} 
                                  & \ding{51}       &                & 87.5       & 95.4       & 96.9      & 74.0     & 85.1     & 87.7    & 62.4    & 78.6   & 83.2   & 10.1     & 17.9    & 22.6    \\
                                  & \ding{51}       & \ding{51}      & \textbf{90.0}       & \textbf{96.4}       & \textbf{97.6}      & \textbf{77.6}     & \textbf{88.0}     & \textbf{90.4}    & \textbf{71.3}    & \textbf{83.7 }  & \textbf{87.3}   & \textbf{20.2 }    & \textbf{33.2}    & \textbf{38.8}    \\ \hline\hline
\multirow{4}{*}{Contrastive}      &                 &                & 83.0       & 93.0       & 95.2      & 72.7     & 82.8     & 85.8    & 53.7    & 67.2   & 74.8   & 8.0      & 13.8    & 17.3    \\
                                  &                 & \ding{51}      & 88.8       & 95.2       & 96.8      & 79.0     & 85.8     & 88.5    & 67.7    & 79.2   & 83.4   & 20.8     & 33.9    & 41.5    \\ \cline{2-15} 
                                  & \ding{51}       &                & 84.5       & 94.0       & 95.9      & 74.6     & 84.7     & 87.8    & 63.4    & 76.9   & 82.5   & 14.6     & 25.2    & 31.2    \\
                                  & \ding{51}       & \ding{51}      & \textbf{90.4}       & \textbf{96.4}       & \textbf{97.6}      & \textbf{79.3}     & \textbf{88.5}     & \textbf{90.7}    & \textbf{73.5}    & \textbf{85.5}   & \textbf{88.9}   & \textbf{31.4}     & \textbf{46.4}    & \textbf{53.5}    \\ \hline\hline
\multirow{4}{*}{Multi-Similarity} &                 &                & 84.0       & 93.3       & 95.5      & 72.7     & 82.7     & 86.5    & 50.7    & 65.1   & 71.5   & 9.4      & 17.9    & 21.7    \\
                                  &                 & \ding{51}      & 89.4       & 96.0       & 97.3      & 81.2     & 89.1     & 90.9    & 64.6    & 76.4   & 80.6   & 18.0     & 30.1    & 36.0    \\ \cline{2-15} 
                                  & \ding{51}       &                & 89.5       & 96.3       & 97.6      & 77.4     & 87.2     & 90.1    & 74.6    & 86.8   & 89.9   & 29.1     & 43.3    & 50.2    \\
                                  & \ding{51}       & \ding{51}      & \textbf{91.5}       & \textbf{97.2}       & \textbf{98.1 }     & \textbf{82.0}     & \textbf{90.4}     & \textbf{91.4}    & \textbf{79.4}    & \textbf{90.6}   & \textbf{93.2}   & \textbf{38.5}     & \textbf{53.9}    & \textbf{60.7}    \\ \hline
\end{tabular}
}
\vspace{6pt}
\caption{\small Ablation. We study the performance gain of three loss functions. For each loss, we train $4$ networks. $2$ of which are baselines (one with Online Hard Mining (OHM) and one without), and the other $2$ are to compare the performance gain introduced by our method (GPM).}
\label{tab:my-table}
\end{table}

\begin{figure}[th]%
\centering
\subfigure[Triplet loss]{\label{fig:valid1}\includegraphics[width=0.32\textwidth]{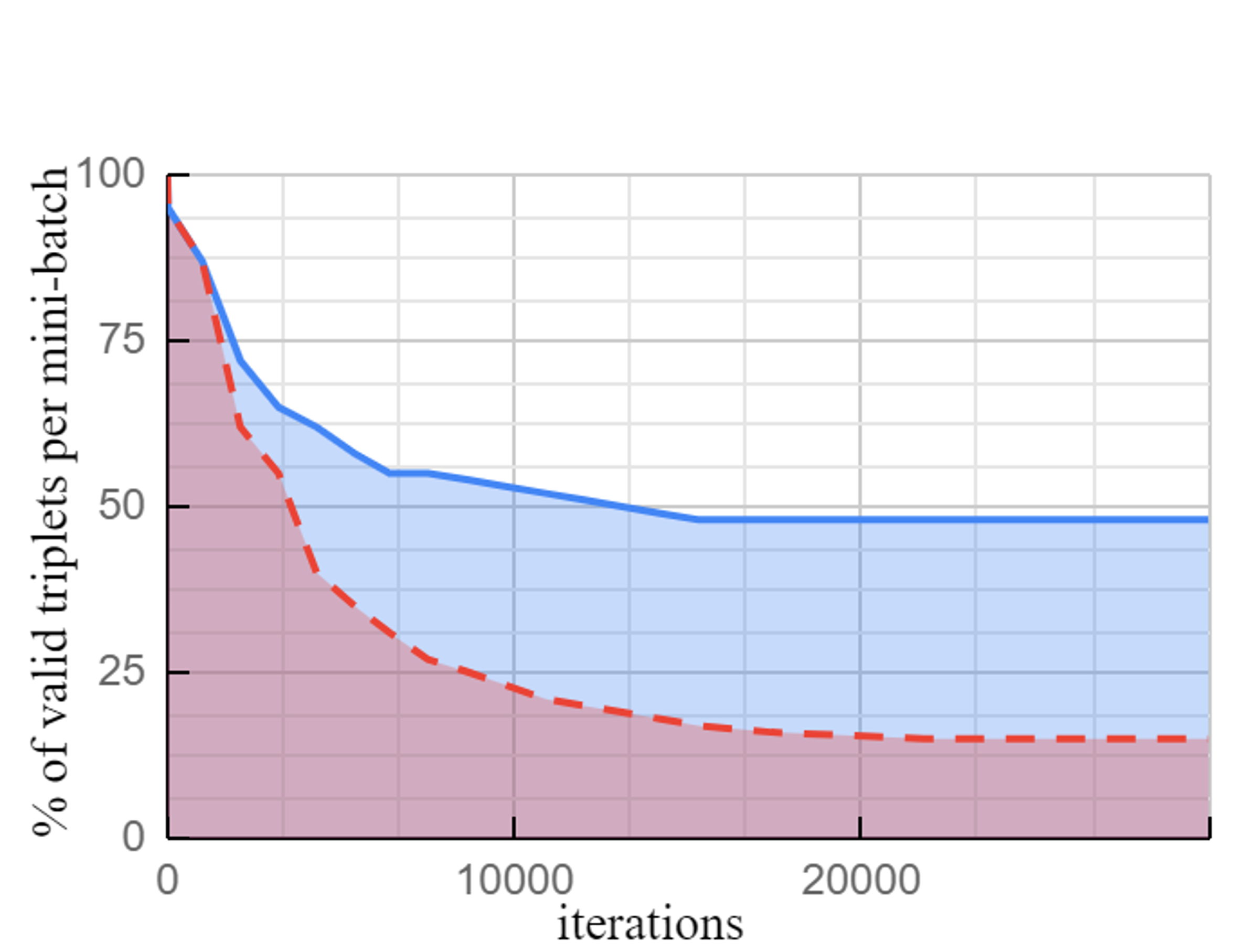}}
\subfigure[Contrastive loss]{\label{fig:valid2}\includegraphics[width=0.32\textwidth]{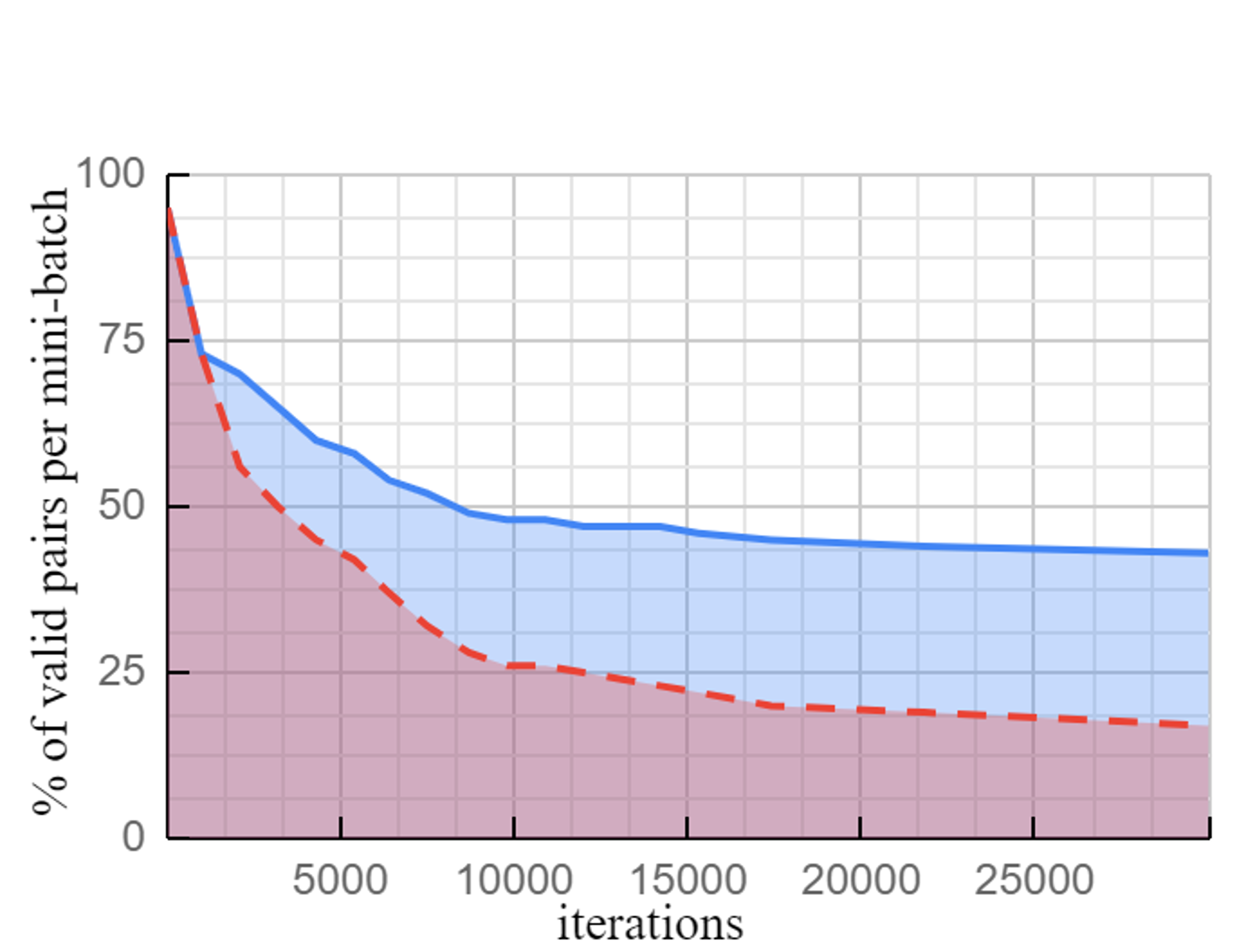}}
\subfigure[Multi-Similarity loss]{\label{fig:valid3}\includegraphics[width=0.32\textwidth]{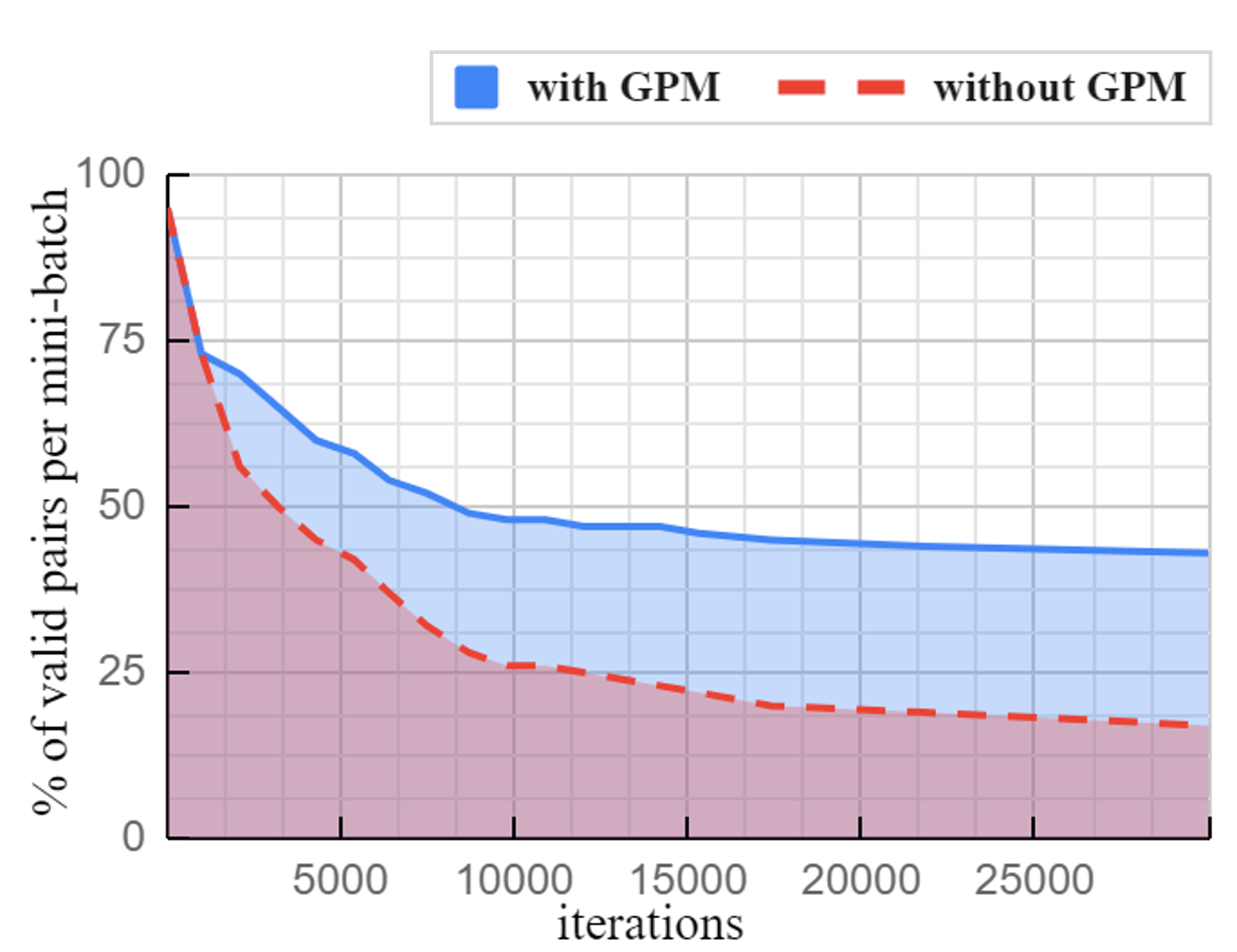}}
\vspace{3pt}
\caption{\small Percentage of valid triplets/pairs per mini-batch during the training. Our technique (GPM) construct highly informative mini-batches, which in turn keeps the number of valid pairs/triplets higher during all the training phase.}
\label{fig:valid_triplets}
\end{figure}

These results make even more sense when we look at the curves on Figure~\ref{fig:valid_triplets} where we keep track of the fraction of informative pairs and triplets within the mini-batch. As training progresses, the network learns to identify most hard samples, making a large fraction of pairs and triplets in the mini-batch uninformative. 
This is highlighted by the red-dotted curve in Figure~\ref{fig:valid_triplets} where the fraction of informative pairs and triplets rapidly decreases to less than $15\%$ after $15$K iterations. More importantly, when we use GPM, where mini-batches are constructed in such a way to incorporate highly informative pairs and triplets, the fraction of informative samples (blue line) stays at around $50\%$ even after $30$K iterations, which explains the performance boost in Table~\ref{tab:my-table}.

\subsection{Mini-batch Size}  
\begin{figure}[h]
  \centering
  \includegraphics[width=0.9\linewidth]{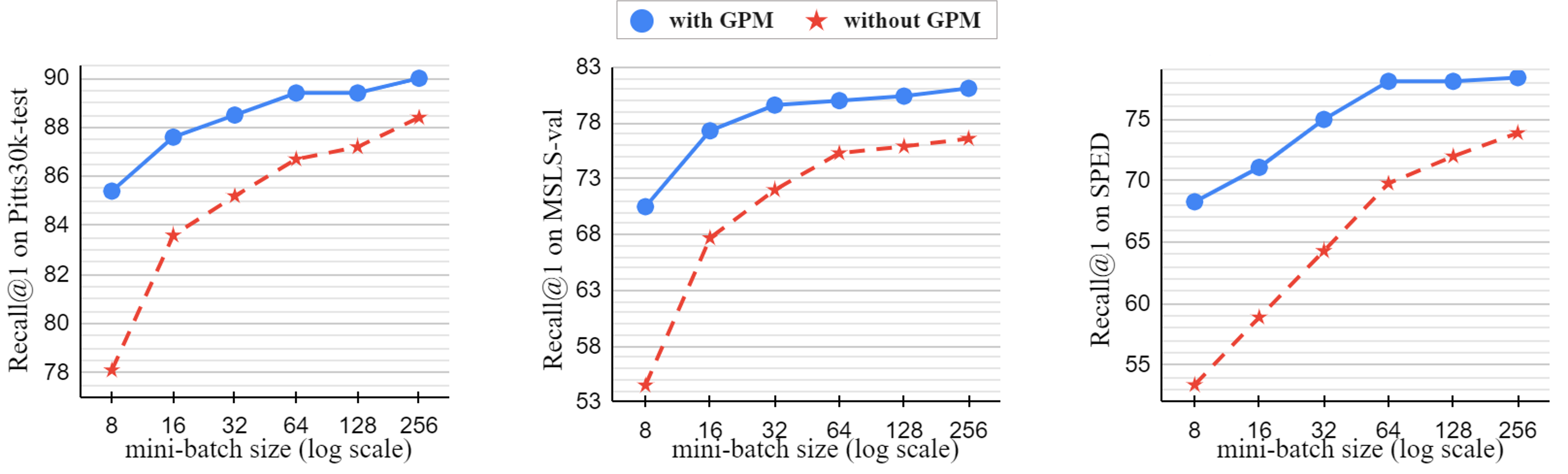}
  \vspace{4pt}
  \caption{\small Impact of the mini-batch size when training with and without GPM. We report recall@1 on Pitts30k-test, MSLS and SPED respectively.  The horizontal axis shows $M$ the number of places in the mini-batch. GPM is effective for a wide range of mini-batch sizes, with more impact when smaller mini-batches are used for training. This is of great importance when training hardware resources are limited.}
\label{fig:batch_size}
\end{figure}

The size of the mini-batch is a key factor in the performance of many pair and triplet based learning approaches. In this experiment, we investigate its impact by using Multi-Similarity loss with and without GPM on three benchmarks. Results are shown in Figure~\ref{fig:batch_size}, where we observe that the smaller the mini-batch size, the lower the performance.
Moreover, when comparing performance with and without GPM, the gap widens as the batch size decreases. This demonstrates that our method brings consistent performance improvements with a wide range of mini-batch sizes. 

\subsection{Memory and computational cost}
Since our method (GPM) requires to add a trainable branch to the network and a memory cache, we investigate the additional computation and memory cost by varying the dimensionality of the proxy head. For each configuration, we train the network for $20$ epochs and record the training time (including the time to build the index and construct mini-batches), the GPU memory required during the training, the size of the memory bank $\Omega$ (Cache size) and the recall@1 performance on Pitts30k-test. 

We first train a baseline model without GPM, and compare against it. Note that for the GPU memory and Cache size, we report the amount of extra memory that was needed compared to the baseline.  
Table~\ref{tab:table2} shows that the baseline model takes $1.93$ hours to finish $20$ training epochs and achieve a recall@1 of $86.6\%$. Since the baseline does not use GPM, there is no extra cache memory (cache size $= 0$).
We then run multiple experiments with GPM, by varying the dimensionality $d'$ of the proxy head (from $32$ to $1024$). The results show that there is  a significant increase in recall@1 performance ($86.6\% \rightarrow 89.4\%$), and a \textit{negligible} amount of GPU and cache memory. For example, by using a proxy of dimension $d'=128$ (as in the above experiments), we end up with $2$MB of extra GPU memory for training $\mathcal{H}$ and $32$MB for the memory cache with \textit{practically} no extra training time. We also notice that proxy with higher dimensionality does not automatically translate to better performance (e.g. GPM with $d'=256$ yields better performance than $d'=1024$).

Particularly, we do another experiment (the rightmost column in table~\ref{tab:table2}) where instead of using a proxy head to generate proxies, we save the NetVLAD representations into cache (we populate $\Omega$ with $32$k-dimensional vectors) and apply global hard mining on them. We end up with $8.0$GB of extra cache memory, more than double the training time and most importantly we get worst recall@1 performance ($88.7\%$ compared to $89.3\%$ when using a $256$-d proxy head). This can be explained by the fact that using the NetVLAD representations resulted in mining the most difficult pairs which is know to impact performance if the dataset contains a certain amount of outliers~\cite{hermans2017defense}. This experiment shows that, even if memory and computation are not a concern, GPM is still a better choice for learning robust representations.

\begin{table}[th]
\centering
\resizebox{\textwidth}{!}{%
\begin{tabular}{l||c||cccccc||c}
                      & \begin{tabular}[c]{@{}c@{}}Baseline\\ (no GPM)\end{tabular} & \multicolumn{6}{c||}{\begin{tabular}[c]{@{}c@{}}Global Proxy-based Hard Mining\\ (GPM)\end{tabular}} & \begin{tabular}[c]{@{}c@{}}Global hard mining \\ without proxy\end{tabular} \\ \hline\hline
Dimensionality        & 0                                                           & 32             & 64             & 128            & 256            & 512            & 1024           & 32768                                                                       \\ \hline
Training time (hours) & 1.93                                                        & 1.93           & 1.93           & 1.93           & 1.94           & 2.05           & 2.1            & 4.83                                                                        \\ \hline
GPU memory (GB)       & 10.4                                                        & +0.002          & +0.002          & +0.002          & +0.03           & +0.06           & +0.14           & +0.0                                                                         \\ \hline
Cache size (GB)       & 0.0                                                         & +0.008          & +0.016          & +0.032          & +0.064          & +0.128          & +0.256          & +8.0                                                                         \\ \hline
Recall@1 (\%)         & 86.6                                                        & 89.1           & 89             & 89.3           & 89.4           & 89             & 89.2           & 88.7                                                                        \\ \hline
\end{tabular}
}
\vspace{7pt}
\caption{\small Memory and computation cost of different dimensions of the proxy head compared against the baseline without GPM). We also compare against global mining without a proxy head, where the memory bank is filled with the highly dimensional NetVLAD representations.}
\label{tab:table2}
\end{table}

\subsection{Qualitative Results}\label{ssec:qualitative}
\begin{figure}[thb]%
\centering
\subfigure[A mini-batch sampled with GPM]{\label{fig:batch1}\includegraphics[width=0.385\textwidth]{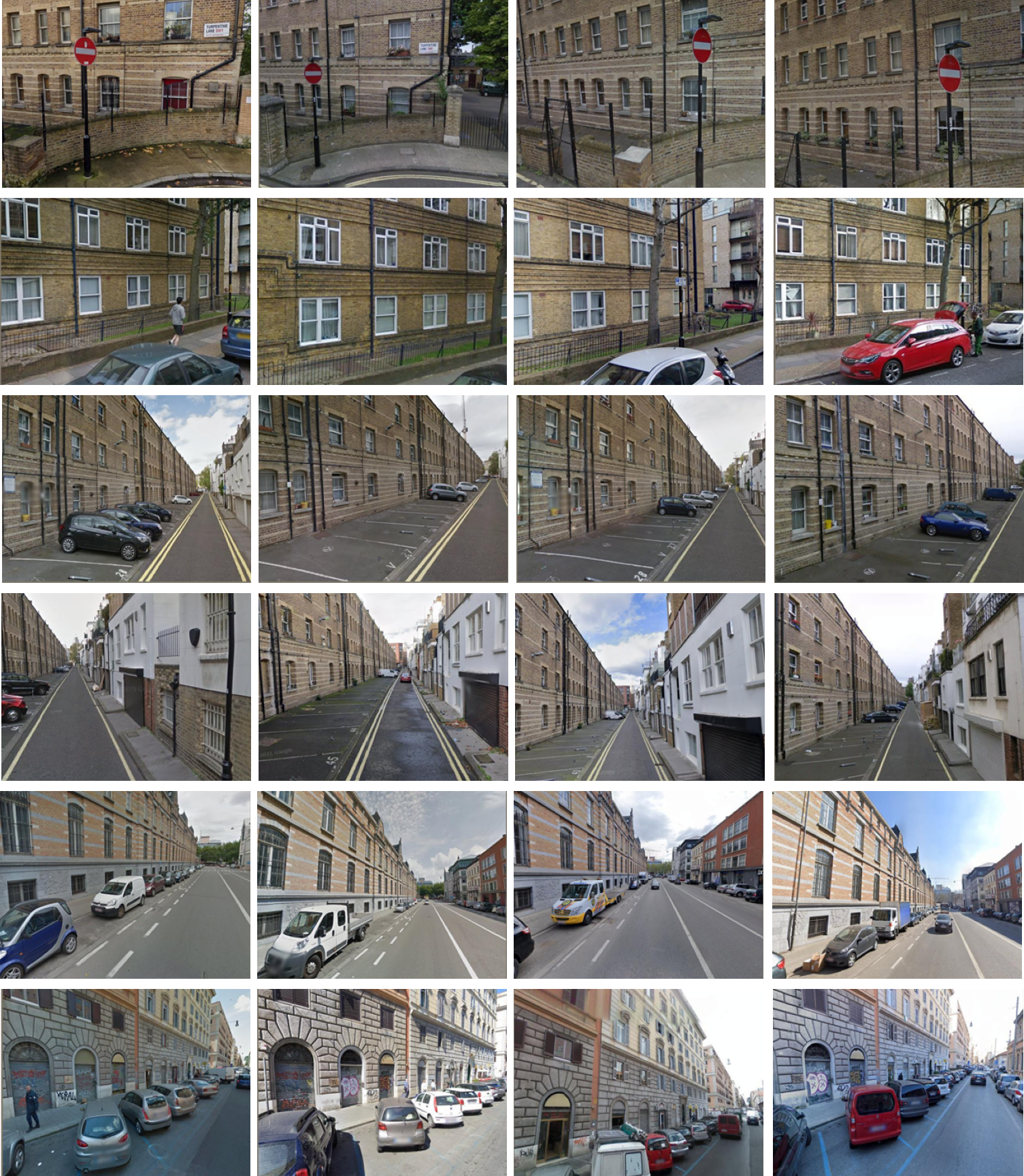}}
\hfill
\subfigure[Valid triplets]{\label{fig:batch2}\includegraphics[width=0.31\textwidth]{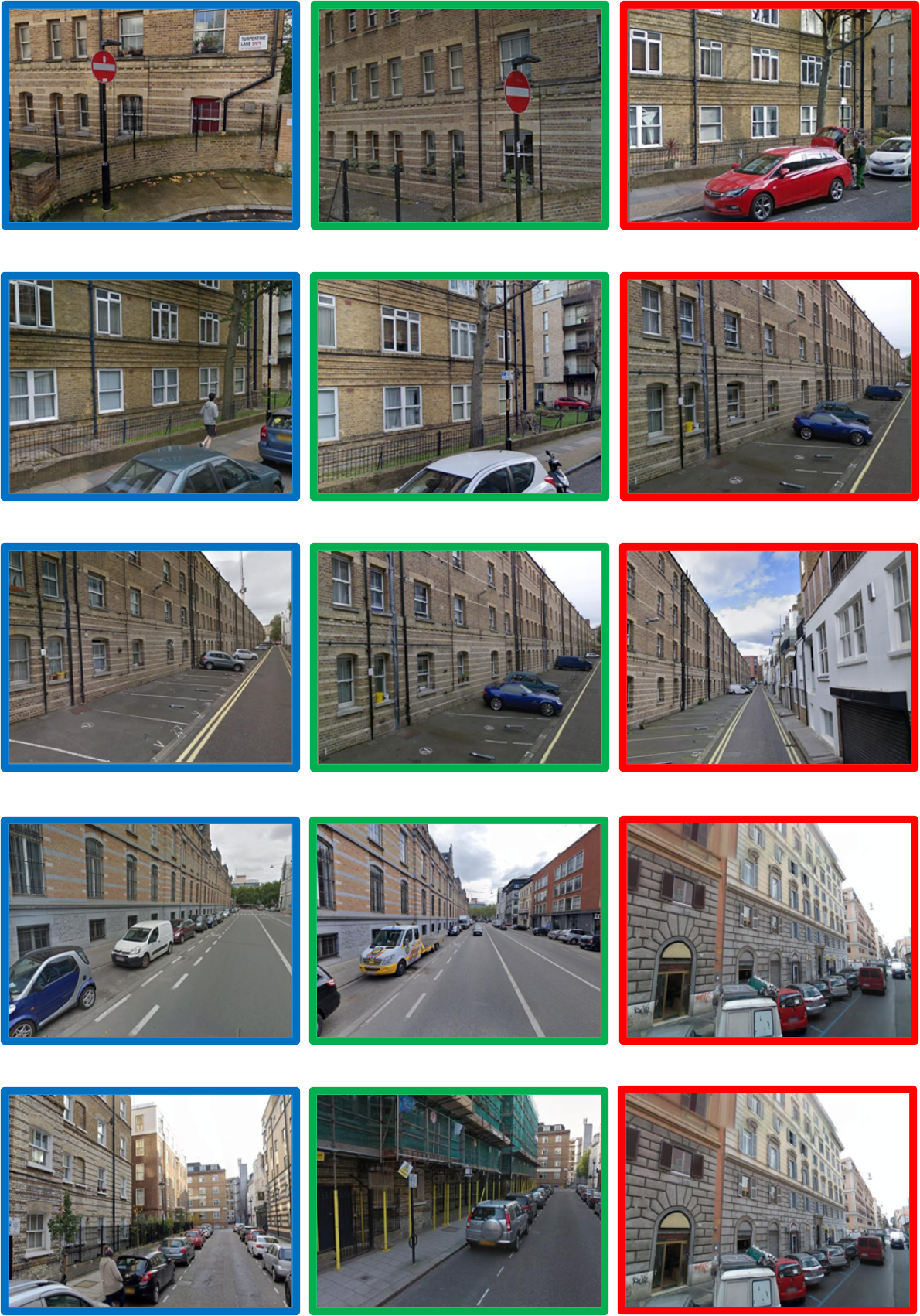}}
\hfill
\subfigure[Valid pairs]{\label{fig:batch3}\includegraphics[width=0.2\textwidth]{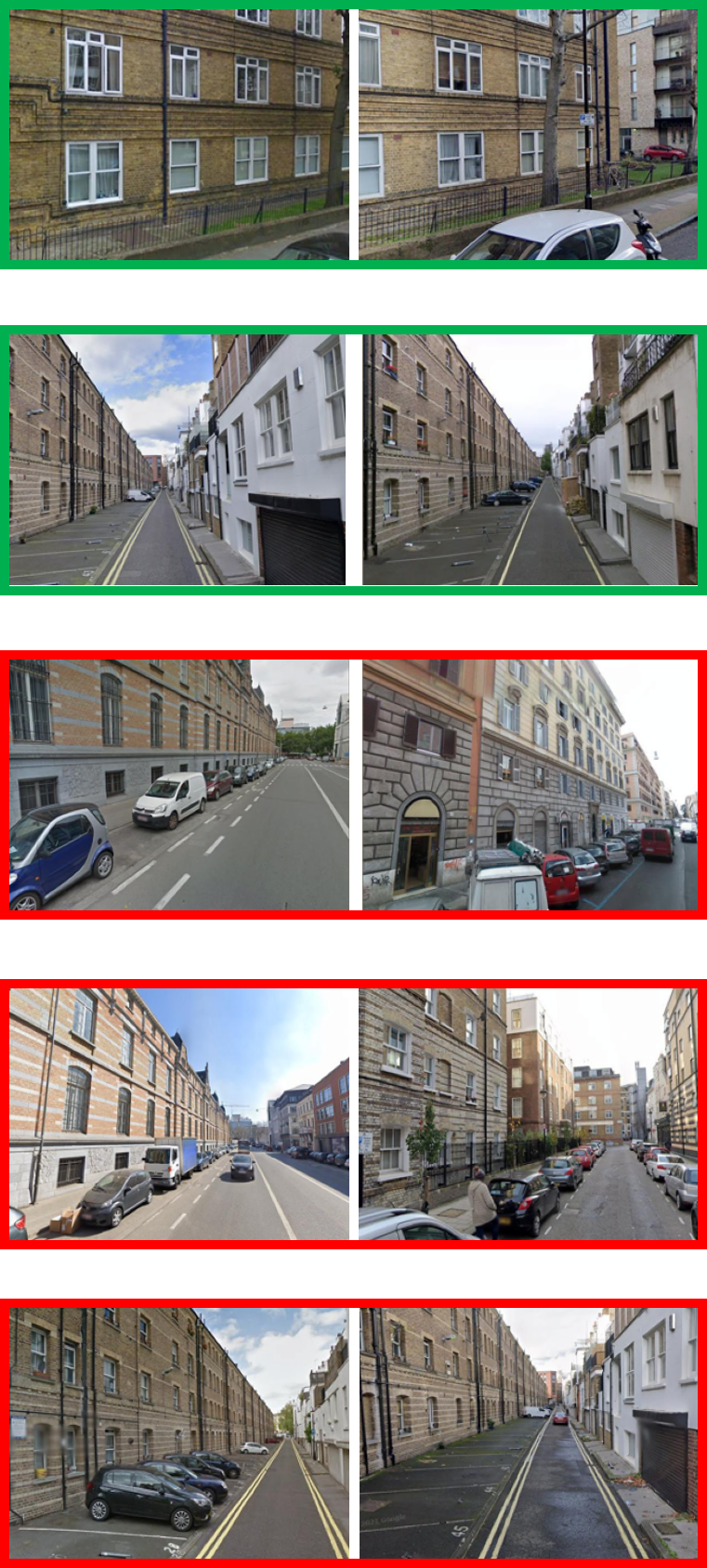}}
\vspace{3pt}
\caption{\small (a) An example of a mini-batch containing $6$ places sampled from a dataset of $65$k places using GPM. Each place is depicted by $4$ images (a row). This highlights the ability of our technique to construct mini-batches containing similar places, which in turn increases the presence of hard pairs and triplets. (b) A subset of hard triplets generated from the mini-batch, each row consists of a triplet with the blue as anchor, green as the positive and red as the hard negative. (c) A subset of positive (green) and negative (red) pairs. All triplets and pairs have been mined in an online fashion from the mini-batch sampled by GPM.}
\label{fig:a_batch}
\end{figure}

Our technique (GPM) relies on the similarity between proxies to form mini-batches comprising visually similar places. In this experiment, we used GPM to sample a mini-batch containing $6$ places ($M=6$) from a database of $65$k different places. Note that the probability of \emph{randomly} sampling $6$ similar places among $65$k is extremely low. We show in Figure~\ref{fig:batch1} a mini-batch of $6$ places sampled using GPM, we notice that all $6$ places are visually similar containing similar textures and structures aligned in a similar manner. In Figures~\ref{fig:batch2}~and~\ref{fig:batch3} we visualize a subset of triplets and pairs mined using OHM on the same mini-batch sampled by GPM. Some triplets contain negatives that are visually extremely difficult to distinguish. This shows how using GPM can ensure, to a certain degree, the presence of visually similar places at each training iteration, increasing the likelihood of hard pairs and triplets, which in turn helps learn robust representations.

\section{Conclusion}
\label{conclusion}
In this paper, we proposed a novel technique that employs compact proxy descriptors to sample highly informative mini-batches at each training iteration with negligible additional memory and computational costs. To do so, we add an auxiliary branch to the baseline network that generates compact place-specific descriptors, which are used to compute one proxy for each place in the dataset. The compactness of these proxies allows to efficiently build a global index that gathers places in the same mini-batch based on the similarity of their proxies.
Our method proved to be very effective in keeping the fraction of informative pairs and triplets at a high level during the entire training phase, resulting in substantial improvement in overall performance. Future works can focus on the architecture of the proxy head and on different ways of building the global index.

\vspace{5pt}
\noindent\textbf{Acknowledgement.} This work has been supported by The Fonds de Recherche du Québec Nature et technologies (FRQNT). We gratefully acknowledge the support of NVIDIA Corporation with the donation of a Quadro RTX 8000 GPU used for our experiments.

\bibliography{bibliography}

\begin{thebibliography}{37}
\providecommand{\natexlab}[1]{#1}
\providecommand{\url}[1]{\texttt{#1}}
\expandafter\ifx\csname urlstyle\endcsname\relax
  \providecommand{\doi}[1]{doi: #1}\else
  \providecommand{\doi}{doi: \begingroup \urlstyle{rm}\Url}\fi

\bibitem[Ali-bey et~al.(2022)Ali-bey, Chaib-draa, and Gigu{\`e}re]{ali2022gsv}
Amar Ali-bey, Brahim Chaib-draa, and Philippe Gigu{\`e}re.
\newblock {\textsc{GSV-Cities}: Toward Appropriate Supervised Visual Place
  Recognition}.
\newblock \emph{Neurocomputing}, 2022.

\bibitem[Arandjelovic et~al.(2016)Arandjelovic, Gronat, Torii, Pajdla, and
  Sivic]{arandjelovic2016netvlad}
Relja Arandjelovic, Petr Gronat, Akihiko Torii, Tomas Pajdla, and Josef Sivic.
\newblock {NetVLAD: CNN architecture for weakly supervised place recognition}.
\newblock In \emph{IEEE Conference on Computer Vision and Pattern Recognition
  (CVPR)}, pages 5297--5307, 2016.

\bibitem[Baik et~al.(2020)Baik, Kim, Shen, Ilg, Lee, and
  Sweeney]{baik2020domain}
Sungyong Baik, Hyo~Jin Kim, Tianwei Shen, Eddy Ilg, Kyoung~Mu Lee, and
  Christopher Sweeney.
\newblock Domain adaptation of learned featuresfor visual localization.
\newblock In \emph{BMVC}, 2020.

\bibitem[Chen et~al.(2021)Chen, Liu, Wang, Bakker, Georgiou, Fieguth, Liu, and
  Lew]{chen2021deep}
Wei Chen, Yu~Liu, Weiping Wang, Erwin Bakker, Theodoros Georgiou, Paul Fieguth,
  Li~Liu, and Michael~S Lew.
\newblock Deep image retrieval: A survey.
\newblock \emph{arXiv preprint arXiv:2101.11282}, 2021.

\bibitem[Chowdhary et~al.(2013)Chowdhary, Johnson, Magree, Wu, and
  Shein]{chowdhary2013gps}
Girish Chowdhary, Eric~N Johnson, Daniel Magree, Allen Wu, and Andy Shein.
\newblock Gps-denied indoor and outdoor monocular vision aided navigation and
  control of unmanned aircraft.
\newblock \emph{Journal of field robotics}, 30\penalty0 (3):\penalty0 415--438,
  2013.

\bibitem[Cieslewski et~al.(2016)Cieslewski, Stumm, Gawel, Bosse, Lynen, and
  Siegwart]{cieslewski2016point}
Titus Cieslewski, Elena Stumm, Abel Gawel, Mike Bosse, Simon Lynen, and Roland
  Siegwart.
\newblock Point cloud descriptors for place recognition using sparse visual
  information.
\newblock In \emph{2016 IEEE International Conference on Robotics and
  Automation (ICRA)}, pages 4830--4836. IEEE, 2016.

\bibitem[Cunningham and Delany(2021)]{cunningham2021k}
Padraig Cunningham and Sarah~Jane Delany.
\newblock k-nearest neighbour classifiers-a tutorial.
\newblock \emph{ACM Computing Surveys (CSUR)}, 54\penalty0 (6):\penalty0 1--25,
  2021.

\bibitem[Engel et~al.(2014)Engel, Sch{\"o}ps, and Cremers]{engel2014lsd}
Jakob Engel, Thomas Sch{\"o}ps, and Daniel Cremers.
\newblock Lsd-slam: Large-scale direct monocular slam.
\newblock In \emph{European conference on computer vision}, pages 834--849.
  Springer, 2014.

\bibitem[Ge(2018)]{ge2018deep}
Weifeng Ge.
\newblock Deep metric learning with hierarchical triplet loss.
\newblock In \emph{Proceedings of the European Conference on Computer Vision
  (ECCV)}, pages 269--285, 2018.

\bibitem[Hadsell et~al.(2006)Hadsell, Chopra, and
  LeCun]{hadsell2006dimensionality}
Raia Hadsell, Sumit Chopra, and Yann LeCun.
\newblock Dimensionality reduction by learning an invariant mapping.
\newblock In \emph{IEEE Conference on Computer Vision and Pattern Recognition
  (CVPR)}, volume~2, pages 1735--1742, 2006.

\bibitem[Hausler et~al.(2021)Hausler, Garg, Xu, Milford, and
  Fischer]{hausler2021patch}
Stephen Hausler, Sourav Garg, Ming Xu, Michael Milford, and Tobias Fischer.
\newblock Patch-netvlad: Multi-scale fusion of locally-global descriptors for
  place recognition.
\newblock In \emph{Proceedings of the IEEE/CVF Conference on Computer Vision
  and Pattern Recognition}, pages 14141--14152, 2021.

\bibitem[He et~al.(2016)He, Zhang, Ren, and Sun]{he2016deep}
Kaiming He, Xiangyu Zhang, Shaoqing Ren, and Jian Sun.
\newblock Deep residual learning for image recognition.
\newblock In \emph{IEEE Conference on Computer Vision and Pattern Recognition
  (CVPR)}, pages 770--778, 2016.

\bibitem[Hermans et~al.(2017)Hermans, Beyer, and Leibe]{hermans2017defense}
Alexander Hermans, Lucas Beyer, and Bastian Leibe.
\newblock In defense of the triplet loss for person re-identification.
\newblock \emph{arXiv preprint arXiv:1703.07737}, 2017.

\bibitem[Kim et~al.(2017)Kim, Dunn, and Frahm]{kim2017learned}
Hyo~Jin Kim, Enrique Dunn, and Jan-Michael Frahm.
\newblock Learned contextual feature reweighting for image geo-localization.
\newblock In \emph{IEEE Conference on Computer Vision and Pattern Recognition
  (CVPR)}, pages 3251--3260, 2017.

\bibitem[Kim et~al.(2020)Kim, Kim, Cho, and Kwak]{kim2020proxy}
Sungyeon Kim, Dongwon Kim, Minsu Cho, and Suha Kwak.
\newblock Proxy anchor loss for deep metric learning.
\newblock In \emph{Proceedings of the IEEE/CVF Conference on Computer Vision
  and Pattern Recognition}, pages 3238--3247, 2020.

\bibitem[Krizhevsky et~al.(2012)Krizhevsky, Sutskever, and
  Hinton]{krizhevsky2012imagenet}
Alex Krizhevsky, Ilya Sutskever, and Geoffrey~E Hinton.
\newblock Imagenet classification with deep convolutional neural networks.
\newblock \emph{Advances in neural information processing systems}, 25, 2012.

\bibitem[Liu et~al.(2020)Liu, Zhang, Hua, and Zhao]{liu2020digging}
Hong Liu, Qian Zhang, Guoliang Hua, and Chenyang Zhao.
\newblock Digging hierarchical information for visual place recognition with
  weighting similarity metric.
\newblock In \emph{2020 IEEE International Conference on Image Processing
  (ICIP)}, pages 1456--1460. IEEE, 2020.

\bibitem[Liu et~al.(2019)Liu, Li, and Dai]{liu2019stochastic}
Liu Liu, Hongdong Li, and Yuchao Dai.
\newblock Stochastic attraction-repulsion embedding for large scale image
  localization.
\newblock In \emph{IEEE/CVF International Conference on Computer Vision
  (ICCV)}, pages 2570--2579, 2019.

\bibitem[Maddern et~al.(2017)Maddern, Pascoe, Linegar, and
  Newman]{maddern20171}
Will Maddern, Geoffrey Pascoe, Chris Linegar, and Paul Newman.
\newblock 1 year, 1000 km: The oxford robotcar dataset.
\newblock \emph{The International Journal of Robotics Research}, 36\penalty0
  (1):\penalty0 3--15, 2017.

\bibitem[Menghani(2021)]{menghani2021efficient}
Gaurav Menghani.
\newblock Efficient deep learning: A survey on making deep learning models
  smaller, faster, and better.
\newblock \emph{arXiv preprint arXiv:2106.08962}, 2021.

\bibitem[Milford and Wyeth(2012)]{milford2012seqslam}
Michael~J Milford and Gordon~F Wyeth.
\newblock Seqslam: Visual route-based navigation for sunny summer days and
  stormy winter nights.
\newblock In \emph{2012 IEEE international conference on robotics and
  automation}, pages 1643--1649. IEEE, 2012.

\bibitem[Musgrave et~al.(2020)Musgrave, Belongie, and Lim]{musgrave2020metric}
Kevin Musgrave, Serge Belongie, and Ser-Nam Lim.
\newblock A metric learning reality check.
\newblock In \emph{European Conference on Computer Vision}, pages 681--699.
  Springer, 2020.

\bibitem[Sattler et~al.(2017)Sattler, Torii, Sivic, Pollefeys, Taira, Okutomi,
  and Pajdla]{sattler2017large}
Torsten Sattler, Akihiko Torii, Josef Sivic, Marc Pollefeys, Hajime Taira,
  Masatoshi Okutomi, and Tomas Pajdla.
\newblock Are large-scale 3d models really necessary for accurate visual
  localization?
\newblock In \emph{Proceedings of the IEEE Conference on Computer Vision and
  Pattern Recognition}, pages 1637--1646, 2017.

\bibitem[Schroff et~al.(2015)Schroff, Kalenichenko, and
  Philbin]{schroff2015facenet}
Florian Schroff, Dmitry Kalenichenko, and James Philbin.
\newblock Facenet: A unified embedding for face recognition and clustering.
\newblock In \emph{Proceedings of the IEEE conference on computer vision and
  pattern recognition}, pages 815--823, 2015.

\bibitem[Seymour et~al.(2019)Seymour, Sikka, Chiu, Samarasekera, and
  Kumar]{seymour2019semantically}
Zachary Seymour, Karan Sikka, Han-Pang Chiu, Supun Samarasekera, and Rakesh
  Kumar.
\newblock Semantically-aware attentive neural embeddings for long-term 2d
  visual localization.
\newblock In \emph{British Machine Vision Conference (BMVC)}, 2019.

\bibitem[Smirnov et~al.(2018)Smirnov, Melnikov, Oleinik, Ivanova, Kalinovskiy,
  and Luckyanets]{smirnov2018hard}
Evgeny Smirnov, Aleksandr Melnikov, Andrei Oleinik, Elizaveta Ivanova, Ilya
  Kalinovskiy, and Eugene Luckyanets.
\newblock Hard example mining with auxiliary embeddings.
\newblock In \emph{Proceedings of the IEEE Conference on Computer Vision and
  Pattern Recognition Workshops}, pages 37--46, 2018.

\bibitem[Thoma et~al.(2020)Thoma, Paudel, and Gool]{thoma2020soft}
Janine Thoma, Danda~Pani Paudel, and Luc~V Gool.
\newblock Soft contrastive learning for visual localization.
\newblock \emph{Advances in Neural Information Processing Systems},
  33:\penalty0 11119--11130, 2020.

\bibitem[Torii et~al.(2013)Torii, Sivic, Pajdla, and Okutomi]{torii2013visual}
Akihiko Torii, Josef Sivic, Tomas Pajdla, and Masatoshi Okutomi.
\newblock Visual place recognition with repetitive structures.
\newblock In \emph{IEEE Conference on Computer Vision and Pattern Recognition
  (CVPR)}, pages 883--890, 2013.

\bibitem[Wang et~al.(2022)Wang, Shen, Zuo, Zhou, and Zheng]{wang2022transvpr}
Ruotong Wang, Yanqing Shen, Weiliang Zuo, Sanping Zhou, and Nanning Zheng.
\newblock Transvpr: Transformer-based place recognition with multi-level
  attention aggregation.
\newblock In \emph{Proceedings of the IEEE/CVF Conference on Computer Vision
  and Pattern Recognition}, pages 13648--13657, 2022.

\bibitem[Wang et~al.(2019)Wang, Han, Huang, Dong, and Scott]{wang2019multi}
Xun Wang, Xintong Han, Weilin Huang, Dengke Dong, and Matthew~R Scott.
\newblock Multi-similarity loss with general pair weighting for deep metric
  learning.
\newblock In \emph{IEEE/CVF Conference on Computer Vision and Pattern
  Recognition (CVPR)}, pages 5022--5030, 2019.

\bibitem[Warburg et~al.(2020)Warburg, Hauberg, L{\'o}pez-Antequera, Gargallo,
  Kuang, and Civera]{warburg2020mapillary}
Frederik Warburg, Soren Hauberg, Manuel L{\'o}pez-Antequera, Pau Gargallo,
  Yubin Kuang, and Javier Civera.
\newblock Mapillary street-level sequences: A dataset for lifelong place
  recognition.
\newblock In \emph{IEEE/CVF Conference on Computer Vision and Pattern
  Recognition (CVPR)}, pages 2626--2635, 2020.

\bibitem[Wu et~al.(2017)Wu, Manmatha, Smola, and Krahenbuhl]{wu2017sampling}
Chao-Yuan Wu, R~Manmatha, Alexander~J Smola, and Philipp Krahenbuhl.
\newblock Sampling matters in deep embedding learning.
\newblock In \emph{Proceedings of the IEEE International Conference on Computer
  Vision}, pages 2840--2848, 2017.

\bibitem[Yang et~al.(2022)Yang, Bastan, Zhu, Gray, and
  Samaras]{yang2022hierarchical}
Zhibo Yang, Muhammet Bastan, Xinliang Zhu, Douglas Gray, and Dimitris Samaras.
\newblock Hierarchical proxy-based loss for deep metric learning.
\newblock In \emph{Proceedings of the IEEE/CVF Winter Conference on
  Applications of Computer Vision}, pages 1859--1868, 2022.

\bibitem[Yao et~al.(2022)Yao, Bai, Zhang, Zhang, Sun, Chen, Li, and
  Yu]{yao2022pcl}
Xufeng Yao, Yang Bai, Xinyun Zhang, Yuechen Zhang, Qi~Sun, Ran Chen, Ruiyu Li,
  and Bei Yu.
\newblock Pcl: Proxy-based contrastive learning for domain generalization.
\newblock In \emph{Proceedings of the IEEE/CVF Conference on Computer Vision
  and Pattern Recognition}, pages 7097--7107, 2022.

\bibitem[Zaffar et~al.(2021)Zaffar, Garg, Milford, Kooij, Flynn,
  McDonald-Maier, and Ehsan]{zaffar2021vpr}
Mubariz Zaffar, Sourav Garg, Michael Milford, Julian Kooij, David Flynn, Klaus
  McDonald-Maier, and Shoaib Ehsan.
\newblock Vpr-bench: An open-source visual place recognition evaluation
  framework with quantifiable viewpoint and appearance change.
\newblock \emph{International Journal of Computer Vision}, pages 1--39, 2021.

\bibitem[Zhang et~al.(2021)Zhang, Wang, and Su]{zhang2021visual}
Xiwu Zhang, Lei Wang, and Yan Su.
\newblock Visual place recognition: A survey from deep learning perspective.
\newblock \emph{Pattern Recognition}, 113:\penalty0 107760, 2021.

\bibitem[Zhu et~al.(2020)Zhu, Li, Wang, and Zhao]{zhu2020regional}
Yingying Zhu, Biao Li, Jiong Wang, and Zhou Zhao.
\newblock Regional relation modeling for visual place recognition.
\newblock In \emph{Proceedings of the 43rd International ACM SIGIR Conference
  on Research and Development in Information Retrieval}, pages 821--830, 2020.

\end{thebibliography}
\end{document}